\documentclass[10pt,twocolumn,letterpaper]{article}

\usepackage{iccv}
\usepackage{times}
\usepackage{epsfig}
\usepackage{graphicx}
\usepackage{amsmath}
\usepackage{amssymb}
\usepackage{booktabs}       
\usepackage{multirow}
\usepackage{pifont}
\DeclareMathOperator*{\argmax}{arg\,max}
\DeclareMathOperator*{\argmin}{arg\,min}
\usepackage{capt-of}
\usepackage{varwidth}
\newsavebox\tmpbox
\usepackage{algorithm, algorithmic}

\usepackage[pagebackref=true,breaklinks=true,letterpaper=true,colorlinks,bookmarks=false]{hyperref}

\iccvfinalcopy 

\ificcvfinal\fi

\begin{document}

\title{AutoFormer: Searching Transformers for Visual Recognition}
\author{Minghao Chen$^{1,*}$, Houwen Peng$^{2,*,\dagger}$, Jianlong Fu$^2$, Haibin Ling$^1$ \\ $^1$Stony Brook University \quad $^2$Microsoft Research Asia
}

\maketitle
\ificcvfinal\thispagestyle{empty}\fi

\begin{abstract}
Recently, pure transformer-based models have shown great potentials for vision tasks such as image classification and detection. However, the design of transformer networks is challenging. It has been observed that the depth, embedding dimension, and number of heads can largely affect the performance of vision transformers. Previous models configure these dimensions based upon manual crafting. In this work, we propose a new one-shot architecture search framework, namely \emph{AutoFormer}, dedicated to vision transformer search. AutoFormer entangles the weights of different blocks in the same layers during supernet training. Benefiting from the strategy, the trained supernet allows thousands of subnets to be very well-trained. Specifically, the performance of these subnets with weights inherited from the supernet is comparable to those retrained from scratch. Besides, the searched models, which we refer to AutoFormers, surpass the recent state-of-the-arts such as ViT and DeiT. In particular, AutoFormer-tiny/small/base achieve 74.7\%/81.7\%/82.4\% top-1 accuracy on ImageNet with 5.7M/22.9M/53.7M parameters, respectively. Lastly, we verify the transferability of AutoFormer by providing the performance on downstream benchmarks and distillation experiments. Code and models are available at \href{https://github.com/microsoft/AutoML}{https://github.com/microsoft/AutoML}.
\end{abstract}

\newcommand\blfootnote[1]{%
\begingroup 
\renewcommand\thefootnote{}\footnote{#1}%
\addtocounter{footnote}{-1}%
\endgroup 
}
{
	\blfootnote{
	 $^*$Equal contributions. Work performed when Minghao is an intern of MSRA. ~ $^\dagger$ Corresponding author: \href{mailto:houwen.peng@microsoft.com}{\color{black}{houwen.peng@microsoft.com}}.
	}
}

\section{Introduction}
\begin{figure}
    \centering
\includegraphics[width=0.95\columnwidth,height=0.7\columnwidth]{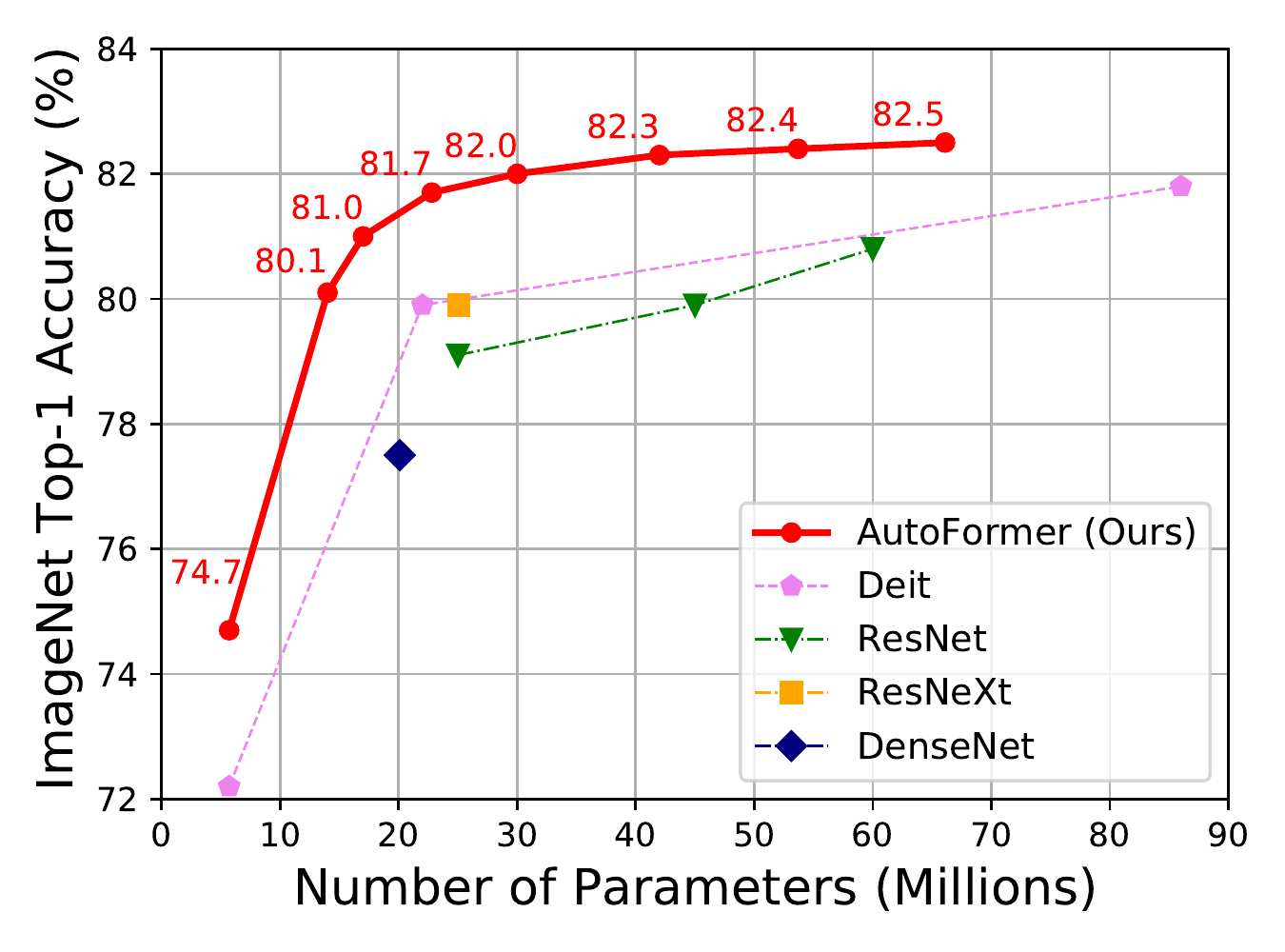}
    \caption{The comparison between AutoFormers and transformer-based, convolution-based and architecture-searched models, such as DeiT \cite{DeiT}, and ResNet \cite{he2016deep}.}
    \label{fig:acc}
    \vspace{-5mm}
\end{figure}
Vision transformer recently has drawn significant attention in computer vision because of its high model capability and superior potentials in capturing long-range dependencies. 
Building on top of transformers~\cite{vaswani2017attention}, modern state-of-the-art models, such as ViT~\cite{dosovitskiy2020vit} and DeiT~\cite{DeiT}, are
able to learn powerful visual representations from images and achieve very competitive performance compared to previous convolutional neural network models~\cite{Survey1, Survey2}.

\begin{figure*}[th]
\centering
\begin{minipage}[t]{0.246\textwidth}
\centering
\includegraphics[width=1\textwidth]{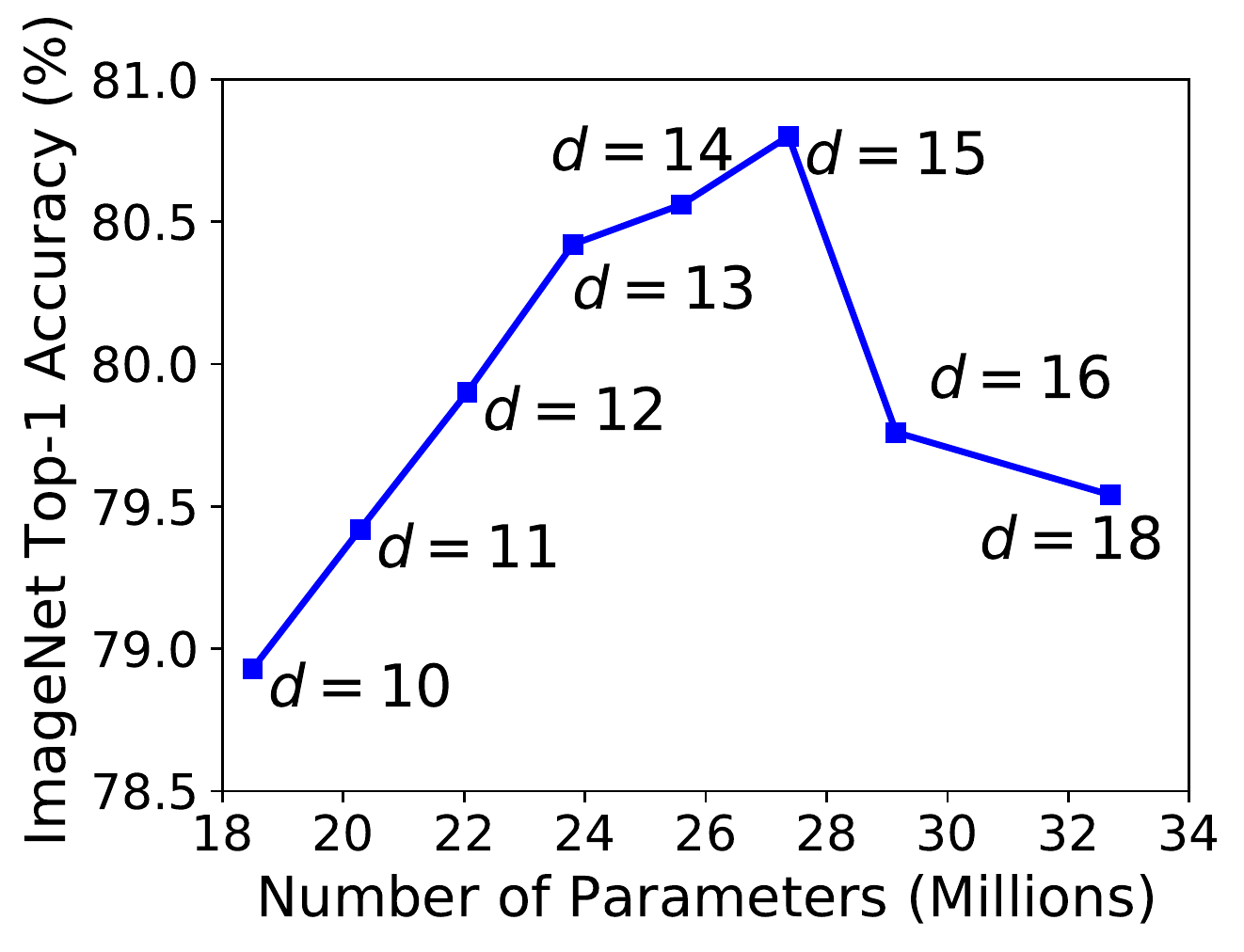}
\end{minipage}
\begin{minipage}[t]{0.246\textwidth}
\centering
\includegraphics[width=1\textwidth]{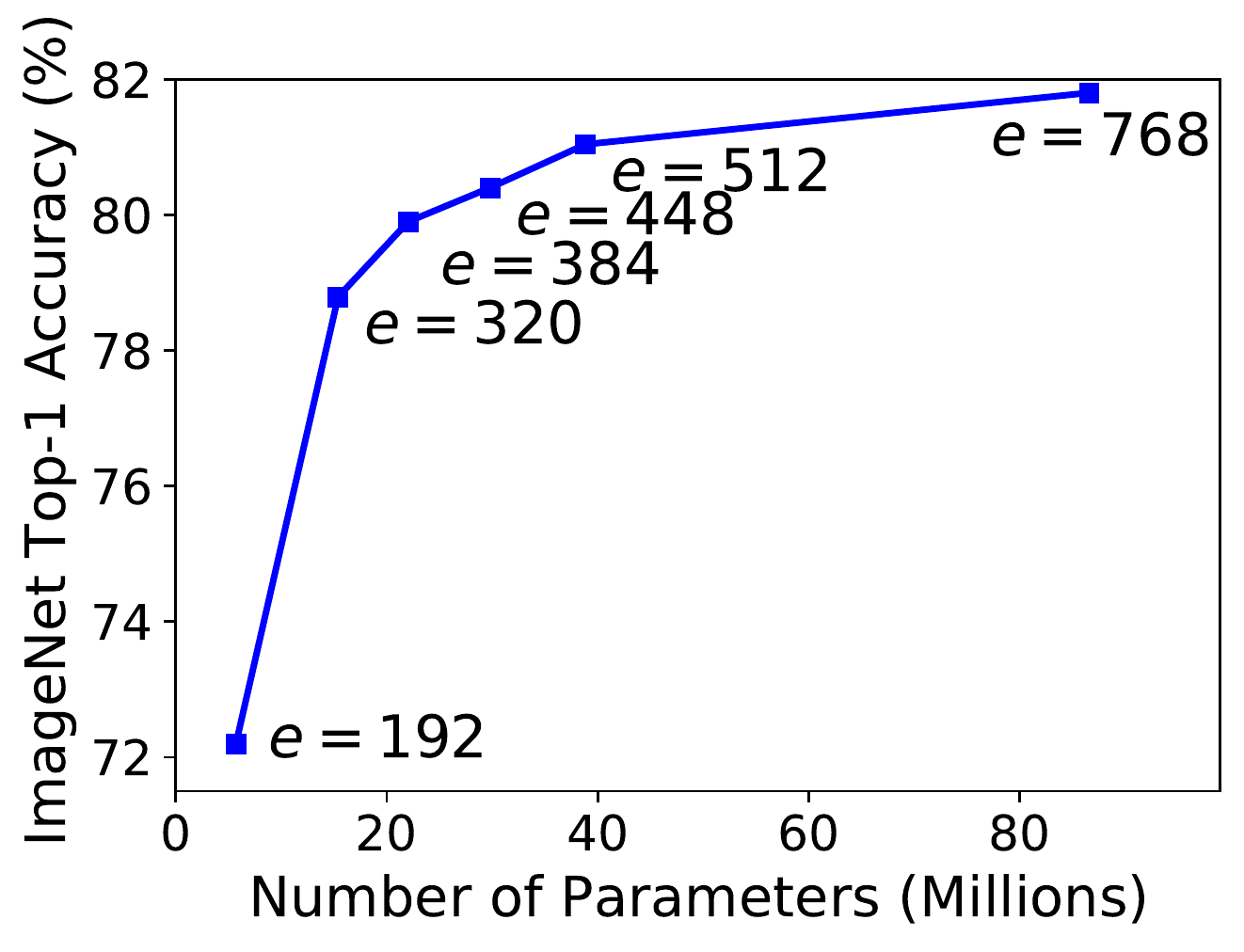}
\end{minipage}
\begin{minipage}[t]{0.246\textwidth}
\centering
\includegraphics[width=1\textwidth]{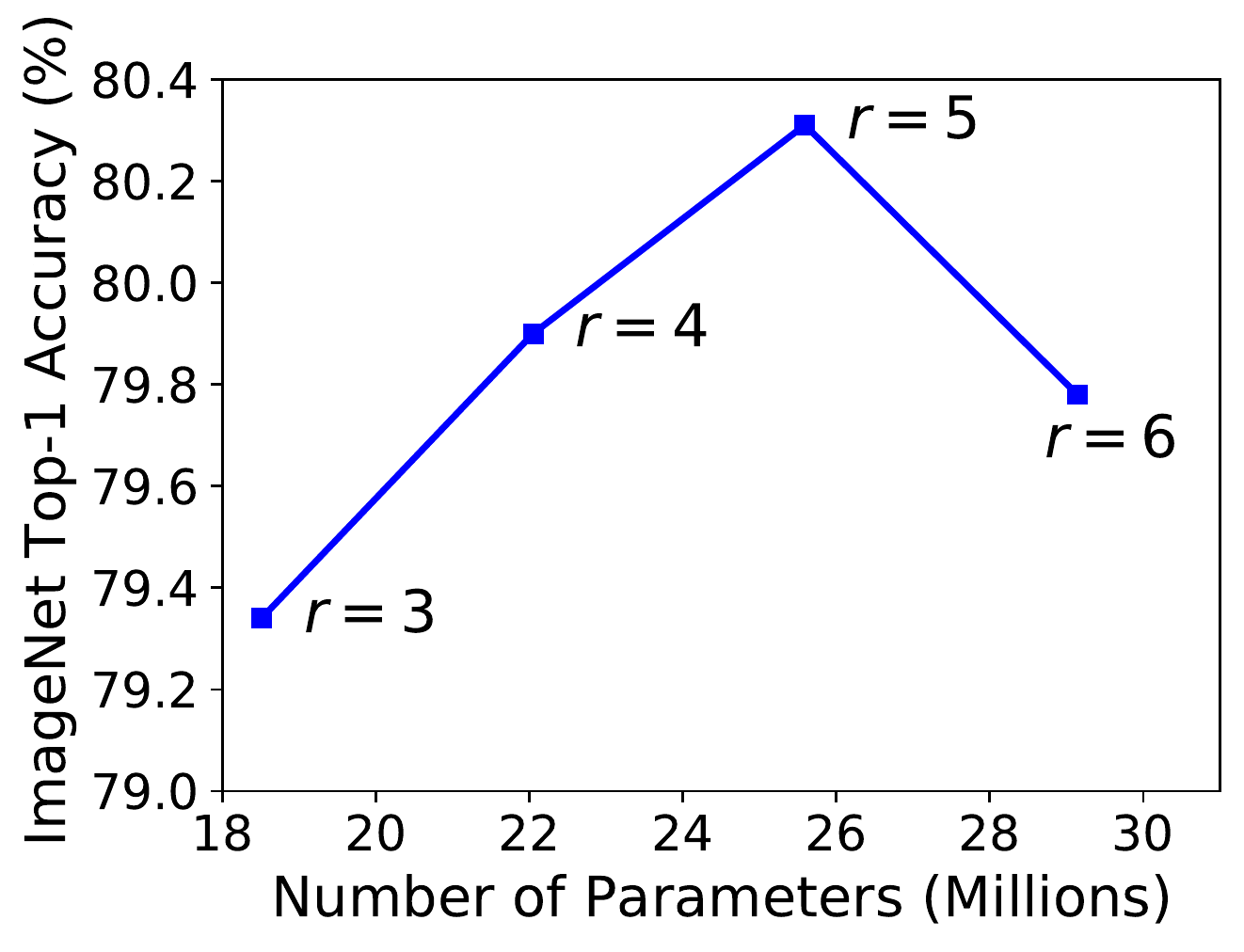}
\end{minipage}
\begin{minipage}[t]{0.246\textwidth}
\centering
\includegraphics[width=1\textwidth]{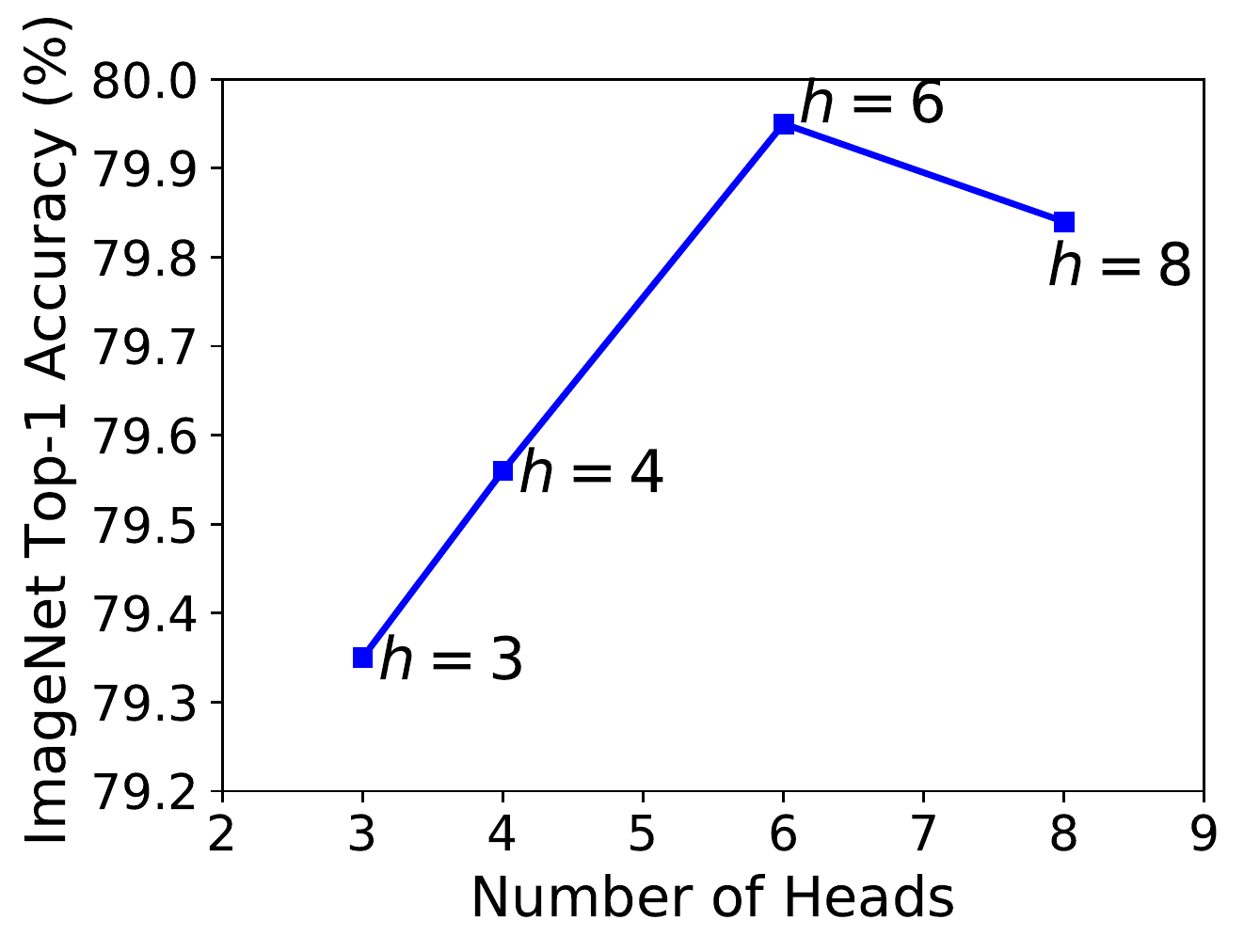}
\end{minipage}
\caption{Adjust a baseline model with different embedding dimension ($e$), depth ($d$), MLP ratio ($r$), and number of heads ($h$) coefficients under the same training recipe, where MLP ratio( the ratio of hidden dimension to the embedding dimension in the multi-layer perceptron). We set the baseline model with $d=12, r=4, e=384, h=6$. \textbf{Note:} number of heads does not affect the model size and complexity if we fix the $Q$-$K$-$V$ dimension.
}

\label{fig:motivation}
\vspace{-4mm}
\end{figure*}

However, the design of transformer neural architectures is nontrivial. For example, how to choose the best network depth, embedding dimension and/or head number? These factors are all critical for elevating model capacity, yet finding a good combination of them is difficult. 
As seen in Fig.~\ref{fig:motivation}, increasing the depth, head number and MLP ratio (the ratio of hidden dimension to the embedding dimension in the multi-layer perceptron) of transformers helps achieve higher accuracy at first but get overfit after hitting the peak value. Scaling up the embedding dimension can improve model capability, but the accuracy gain plateaus for larger models. These phenomenons demonstrate the challenge of designing optimal transformer architectures.

Previous works on designing vision transformers are based upon manual crafting, which heavily relies on human expertise and typically requires a deal of trial-and-error~\cite{dosovitskiy2020vit,DeiT,T2TViT}. There are a few works on automating transformer design using neural architecture search (NAS)~\cite{EvolvedTrans,HAT}. 
However, they are all concentrated on natural language tasks, such as machine translation, 
which are quite different from computer vision tasks. As a result, it is hard to generalize prior automatic search algorithms to find effective vision transformer architectures.  

In this work, we present a new architecture search algorithm, named \emph{AutoFormer}, dedicated to finding pure vision transformer models. Our approach mainly addresses two challenges in transformer search. 1) How to strike a good combination of the key factors in transformers, such as network depth, embedding dimension and head number? 2)
How to efficiently find out various transformer models that fit different resource constraints and application scenarios?

To tackle the challenges, we construct a large search space covering the main changeable dimensions of transformer, including embedding dimension, number of heads, query/key/value dimension, MLP ratio, and network depth. This space contains a vast number of transformers with diverse structures and model complexities. In particular, it allows the construction of transformers to use different structures of building blocks, thus breaking the convention that all blocks share an identical structure in transformer design. 

To address the efficiency issue, inspired by BigNAS \cite{bignas} and slimmable networks \cite{bignas, yu2018slimmable}, we propose a supernet training strategy called \emph{weight entanglement} dedicated to transformer architecture. 
The central idea is to enable different transformer blocks to share weights for their common parts in each layer. An update of weights in one block will affect all other ones as a whole, such that the weights of different blocks are maximally entangled during training. This strategy is different from most one-shot NAS methods \cite{spos, fairnas, wu2019fbnet}, in which the weights of different blocks are independent for the same layer, as visualized in Fig.~\ref{fig:Difference}. 

We observe a surprising phenomenon when using the proposed weight entanglement for transformer supernet training: \emph{it allows a large number of subnets in the supernet to be very well-trained, such that the performance of these subnets with weights inherited from the supernet are comparable to those retrained from scratch.}
This advantage allows our method to obtain thousands of architectures that can meet different resource constraints while maintaining the same level of accuracy as training from scratch independently. We give a detailed discussion in Section \ref{discussion} exploring the underlying reasons of weight entanglement.

We perform a evolutionary search with a model size constraint over the well-trained supernets to find promising transformers.
Experiments on ImageNet \cite{deng2009imagenet} demonstrate that our method achieves superior performance to the handcrafted state-of-the-art transformer models. For instance, as shown in Fig.~\ref{fig:acc}, with 22.9M parameters, Autoformer achieves a top-1 accuracy of 81.7\%, being 1.8\% and 2.9\% better than DeiT-S~\cite{DeiT} and ViT-S/16~\cite{dosovitskiy2020vit}, respectively. In addition, when transferred to downstream vision classification datasets, our AutoFormer also performs well with fewer parameters, achieving better or comparable results to the best convolutional models, such as EfficientNet~\cite{EfficientNet}.

In summary, we make three major contributions in this paper. 1) To our best knowledge, this work is the first effort to design an automatic search algorithm for finding vision transformer models. 2) We propose a simple yet effective framework for efficient training of transformer supernets. Without extra finetuning or retraining, the trained supernet is able to produce thousands of high quality transformers by inheriting weights from it directly. Such merit allows our method to search diverse models to fit different resource constraints. 3) Our searched models, \ie, AutoFormers, achieve the state-of-the-art results on ImageNet among the vision transformers, and demonstrate promising transferability on downstream tasks. 

\begin{figure*}[!t]
    \centering
    \includegraphics[width=0.9\textwidth]{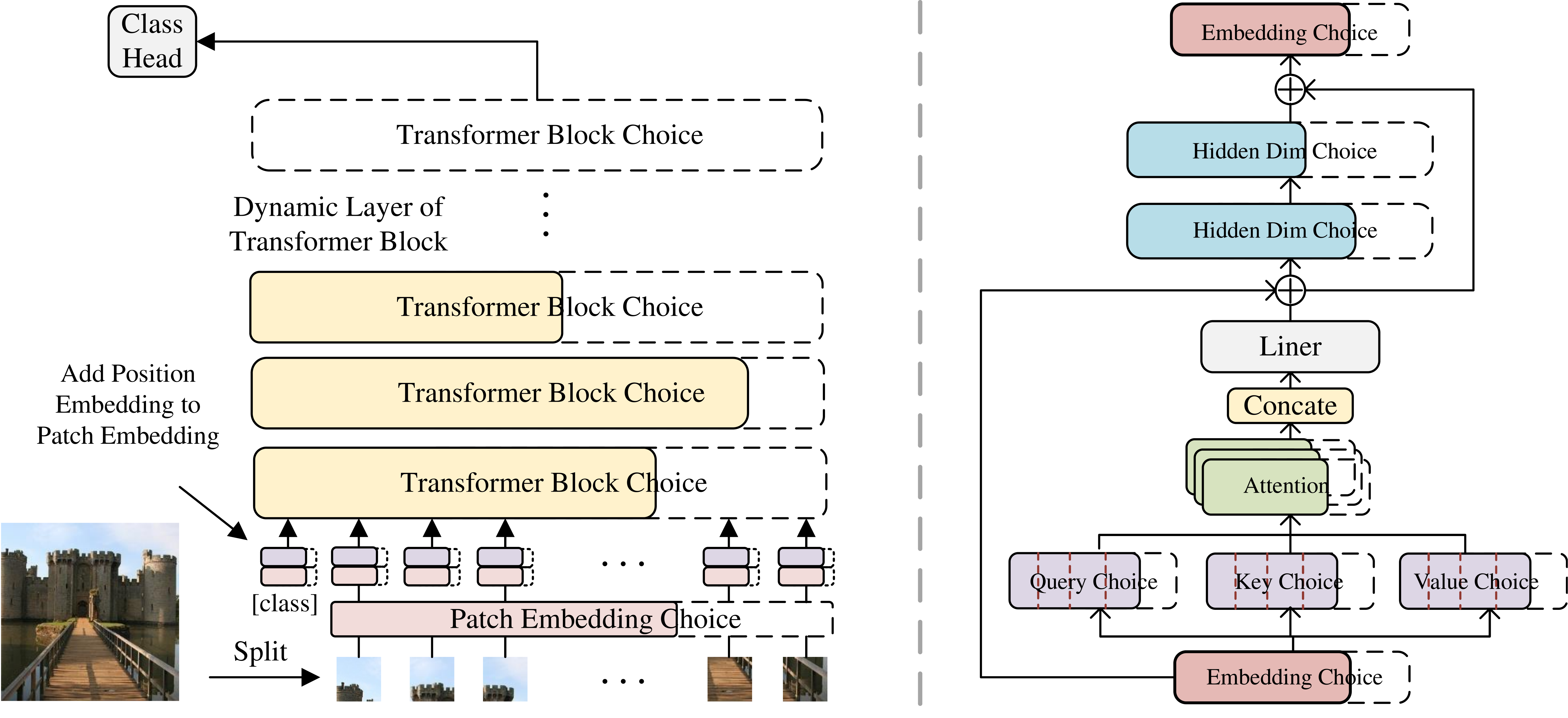}
    \caption{\textbf{Left:}  The overall architecture of the AutoFormer supernet. Note that transformer blocks in each layer and depth are dynamic. The parts in solid lines mean they are chosen while those in dashed lines are not. \textbf{Right:} The detailed transformer block in an AutoFormer. We search for the best block of optimal embedding dimension, number of heads, MLP ratio, $Q$-$K$-$V$ dim in a layer. For more details about the search space, please refer to section \ref{search space}.}
    \label{fig:overview}
    \vspace{-5mm}
\end{figure*}

\section{Background}
Before presenting our method, we first briefly review the background of the vision transformer and one-shot NAS.

\subsection{Vision Transformer}

Transformer is originally designed for natural language tasks \cite{vaswani2017attention, Albert, devlin2019bert}. Recent works, such as ViT and DeiT \cite{dosovitskiy2020vit, DeiT}, show its great potential for visual recognition. In the following, we give the basic pipeline of the vision transformer, which serves as a base architecture of AutoFormer.

Given a 2D image, we first uniformly split it into a sequence of 2D patches just like tokens in natural language processing. We then flatten and transform the patches to $D$-dimension vectors, named patch embeddings, by either linear projection \cite{dosovitskiy2020vit} or several CNN layers \cite{T2TViT}. A learnable [class] embedding is injected into the head of the sequence to represent the whole picture. Position embeddings are added to the patch embeddings to retain positional information. The combined embeddings are then 
fed to a \textit{transformer encoder} described below. At last, a linear layer is used for the final classification.

A transformer encoder consists of alternating blocks of \textit{multihead self-attention} (MSA) and \textit{multi-layer perceptron} (MLP) blocks. LayerNorm (LN) \cite{layernorm} is applied before every block, and residual connections after every block. The details of MSA and MLP are given below.

\textit{Multihead Self-Attention (MSA).} 
In a standard self-attention module, the input sequence $z \in \mathbb{R}^{N \times D}$ will be first linearly transformed to queries $Q \in \mathbb{R}^{N \times D_h}$, keys $K \in \mathbb{R}^{N \times D_h}$ and values $V \in \mathbb{R}^{N \times D_h}$, where $N$ is the number of tokens, $D$ is the embedding dimension, $D_h$ is the $Q$-$K$-$V$ dimension. Then we compute the weighted sum over all values for each element in the sequence. The weights or attention are based on the pairwise similarity between two elements of the sequence:
	\begin{equation}
	\mathrm{Attention} (Q, K, V) = \mathrm{softmax}\Big(\frac{QK^T}{\sqrt{d_h}}\Big)V,
	\label{SA}
	\end{equation}
where $\frac{1}{\sqrt{d_h}}$ is the scaling factor.
Lastly, a fully connected layer is applied. Multihead self-attention splits the queries, keys and values into different heads and performs self-attention in parallel and projects their concatenated outputs.

\textit{Multi-Layer Perceptron (MLP).} The MLP block consists two fully connected layers with an activation function, usually GELU \cite{hendrycks2016gaussian}. In this work, we focus on finding optimal choices of the MLP ratios in each layer.

\subsection{One-Shot NAS}
\label{review}
One-shot NAS typically adopts a weight sharing strategy to avoid training each subnet from scratch~\cite{spos, Cream}. The architecture search space $\mathcal{A}$ is encoded in a supernet, denoted as $\mathcal{N}(\mathcal{A}, W)$, where $W$ is the weight of the supernet. $W$ is shared across all the architecture candidates, \ie, subnets ${\alpha\in \mathcal{A}}$ in $\mathcal{N}$. 
The search of the optimal architecture $\alpha^{*}$ in one-shot NAS is usually formulated as a two-stage optimization problem. The first-stage is to optimize the weight $W$ by 
\begin{equation}
    W_\mathcal{A} = \mathop{ \argmin_{W} } \mathcal{L}_{\emph{train}}(\mathcal{N}(\mathcal{A}, W)),
	\label{eq1}
\end{equation}
where $\mathcal{L}_\emph{train}$ represents the loss function on the training dataset. To reduce memory usage, one-shot methods usually sample subnets from $\mathcal{N}$ for optimization. The second-stage is to search architectures via ranking the performance of subnets ${\alpha\in \mathcal{A}}$ based on the learned weights in $W_\mathcal{A}$:
\begin{equation}
	\alpha^* = \mathop{\argmax}_{\alpha\in \mathcal{A}} \  \mathrm{Acc}_\emph{val} \left( \mathcal{N}(\alpha, w) \right), 
	\label{eq2}
\end{equation}
where the sampled subnet $\alpha$ inherits weight $w$ from $W_\mathcal{A}$, and $\mathrm{Acc}_\emph{val}$ indicates the top-1 accuracy of the architecture $\alpha$ on the validation dataset. Since it is impossible to enumerate all the architectures ${\alpha\in \mathcal{A}}$ for evaluation, prior works resort to random search~\cite{randomNAS,understand_os}, evolution algorithms~\cite{real2019regularized,spos} or reinforcement learning~\cite{ENAS,tan2019mnasnet} to find the most promising one.%

\begin{figure}[!t]
\centering
\begin{minipage}[t]{0.49\columnwidth}
\centering
\includegraphics[width=1\textwidth]{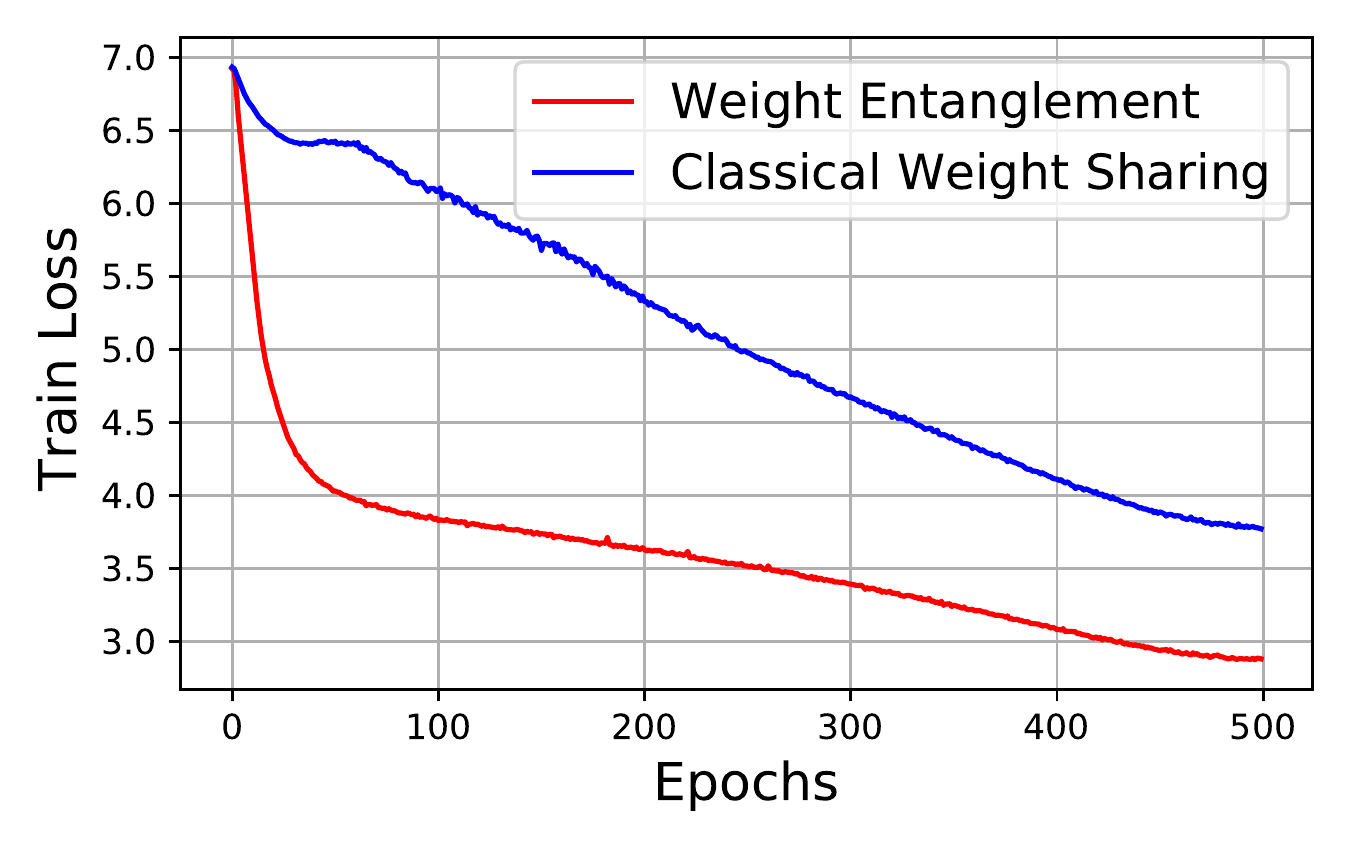} 
\end{minipage}
\begin{minipage}[t]{0.49\columnwidth}
\centering
\includegraphics[width=1\textwidth]{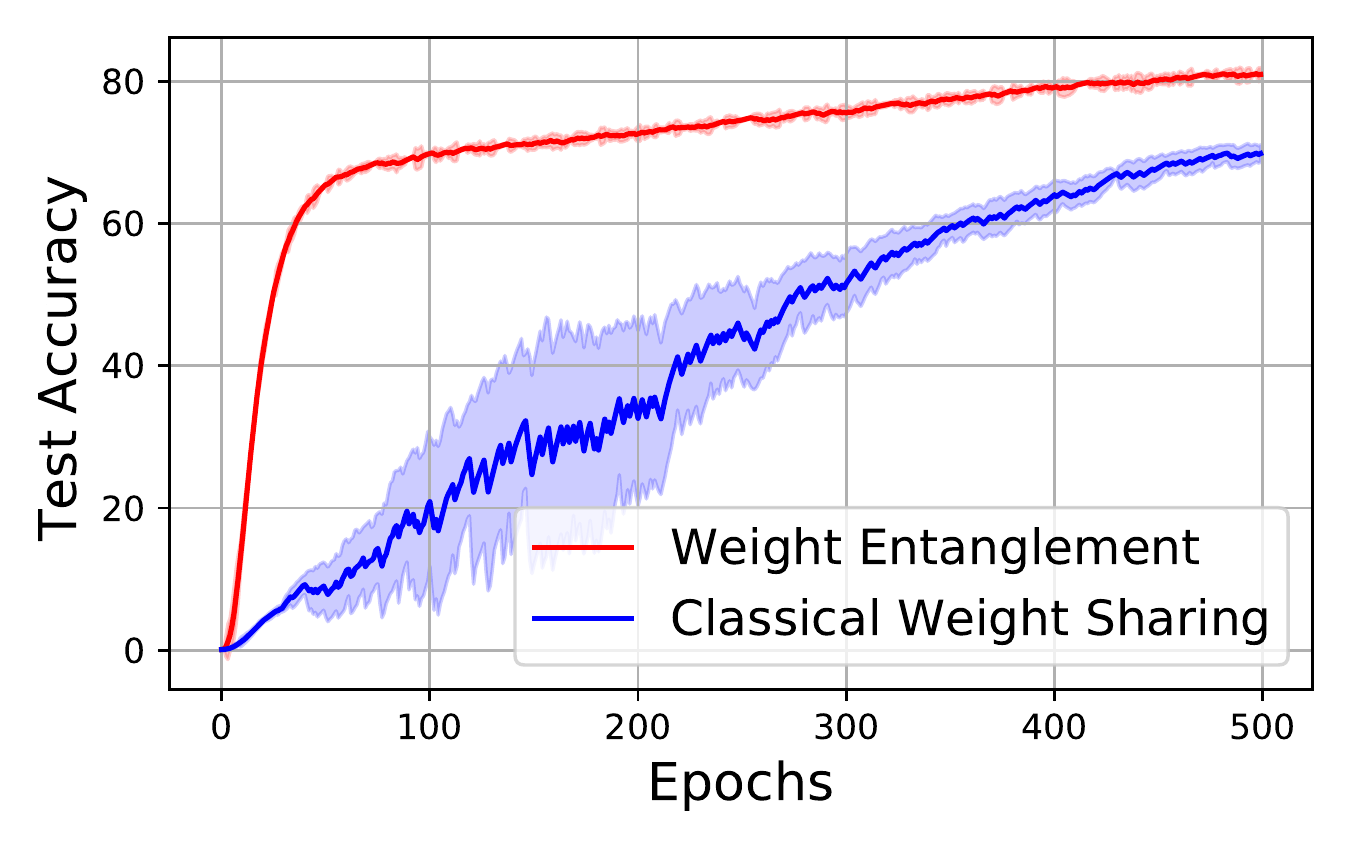} 
\end{minipage}
\caption{\textbf{Left:} Comparison of training loss of supernet between weight entanglement and {classical weight sharing on ImageNet}. \textbf{Right:} Comparison of Top-1 Accuracy on ImageNet of subnets between weight entanglement and classical weight sharing during supernet training.}
\label{Fig:SPOS training}
\vspace{-3mm}
\end{figure}

\section{AutoFormer}

In this section, we first demonstrate that it is impractical to directly apply one-shot NAS for transformer search following classical weight sharing strategy \cite{spos}, using different weights for different blocks in each layer, because of the slow convergence and unsatisfactory performance. Then we propose the weight entanglement strategy for vision transformer to address the issues.
Finally, we present the search space and search pipeline.

\subsection{One-Shot NAS with Weight Entanglement}
\label{Weight Entanglement}

\begin{figure}[!t]
    \centering
    \includegraphics[width=0.45\textwidth]{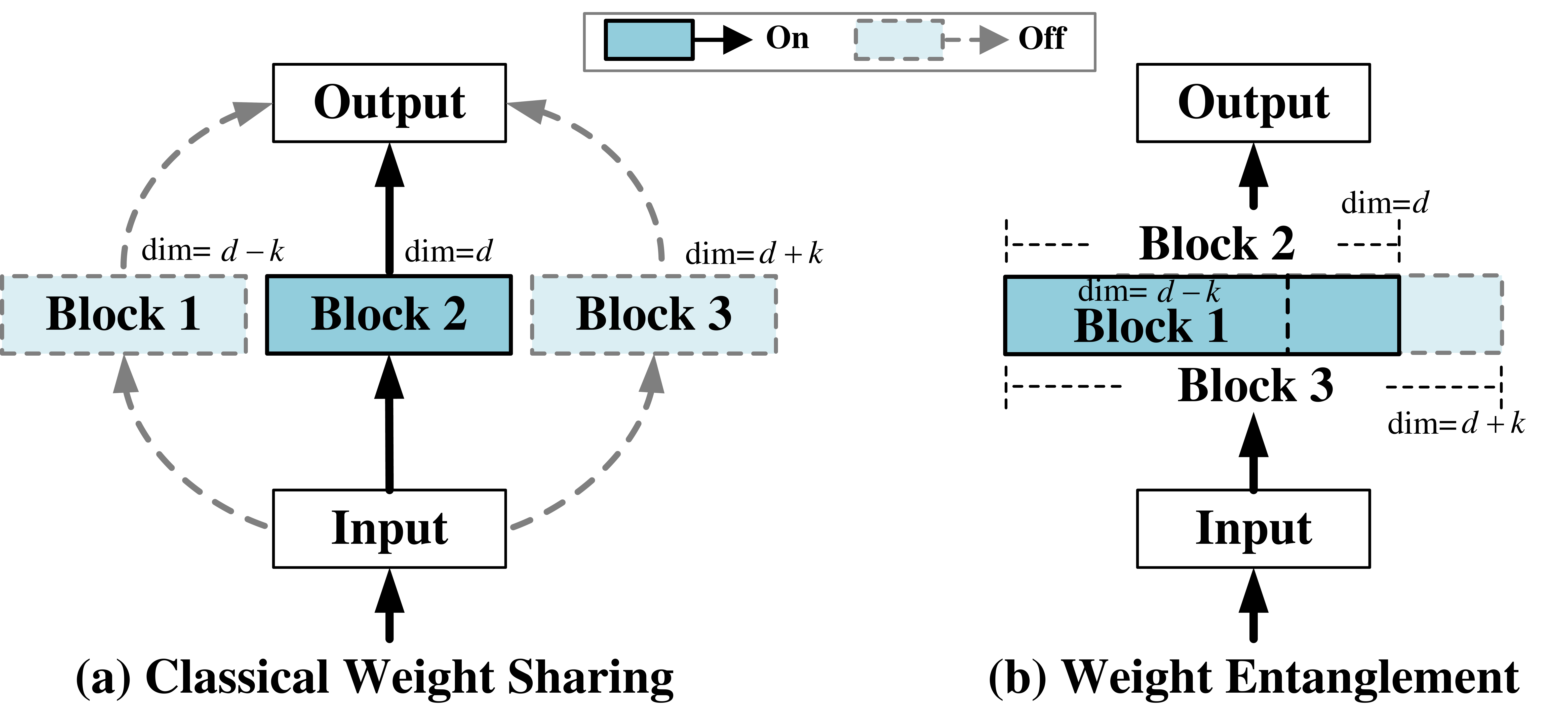}
    \caption{Classical weight sharing versus weight entanglement.}
    \label{fig:Difference}
    \vspace{-4.6mm}

\end{figure}

Prior one-shot NAS methods commonly share weights across architectures during supernet training while decoupling the weights of different operators at the same layer. 
Such strategy performs well when used to search architectures over convolutional neural networks space~\cite{spos, ofa, fairnas, mobilenetv3, EfficientNet}. 
However, in transformer search space, this classical strategy encounters difficulties.
1) Slow convergence. As shown in the Fig.~\ref{Fig:SPOS training} (left), the training loss of the supernet converges slowly. 
The reason might be that the independent training of transformer blocks results in the weights being updated by limited times. 
2) Low performance. The performances of subnets inheriting weights from the one-shot supernet, trained under classical weight sharing strategy, are far below their true performances of training from scratch (see the right part of Fig.~\ref{Fig:SPOS training}). This limits the ranking capacities of supernet. Furthermore, after the search, it is still necessary to perform additional retraining for the searched architectures since the weights are not fully optimized. 
Inspried by BigNAS \cite{bignas} and slimmable networks \cite{yu2018slimmable, yu2019universally},
we propose the weight entanglement training strategy dedicated to vision transformer architecture search. The central idea is to enable different transformer blocks to share weights for their common parts in each layer. More concretely, for a subnet $\alpha\in\mathcal{A}$ with a stack of $l$ layers, we represent its structure and weights as 
\begin{equation}
 \left\{
\begin{aligned}
	\alpha & = (\alpha^{(1)}, ... \alpha^{(i)}, ... \alpha^{(l)}), \\
	w & = (w^{(1)}, ... w^{(i)}, ... w^{(l)}), \\
\end{aligned}
\right.
\end{equation}
where $\alpha^{(i)}$ denotes the sampled block in the $i$-th layer and $w^{(i)}$ is the block weights. During architecture search, there are multiple choices of blocks in each layer. Hence, $\alpha^{(i)}$ and $w^{(i)}$ are actually selected from a set of $n$ block candidates belonging to the search space, which is formulated as
\begin{equation}
 \left\{
\begin{aligned}
	\alpha^{(i)} & \in \{ b^{(i)}_1, ... b^{(i)}_j, ... b^{(i)}_n \}, \\
	w^{(i)} & \in \{ w^{(i)}_{1}, ... w^{(i)}_{j}, ... w^{(i)}_{n} \}, \\
\end{aligned}
\right.
\end{equation}
where $b^{(i)}_j$ is a candidate block in the search space and $w^{(i)}_{j}$ is its weights. 
The weight entanglement strategy enforces that different candidate blocks in the same layer to share as many weights as possible. This requires that, for any two blocks $b^{(i)}_j$ and $b^{(i)}_k$ in the same layer, we have
\begin{equation}
w^{(i)}_{j} \subseteq w^{(i)}_{k} \ \ \mathrm{or} \ \ 
w^{(i)}_{k} \subseteq w^{(i)}_{j}.
\end{equation}
Such within layer weight sharing makes the weight updates of $w^{(i)}_{j}$ and $w^{(i)}_{k}$ entangled with each other. The training of any block will affect the weights of others for their intersected portion, as demonstrated in Fig.~\ref{fig:Difference}. This is different from the classical weight sharing strategy in one-shot NAS, where the building blocks in the same layer are isolated. In other words, in classical weight sharing, for any two blocks $b^{(i)}_j$ and $b^{(i)}_k$, we have $w^{(i)}_{j} \cap w^{(i)}_{k} = \emptyset$.

Note that the proposed weight entanglement strategy is dedicated to work on homogeneous building blocks, such as self-attention modules with different numbers of heads, and multi-layer perceptron with different hidden dimensions.
The underlying reason is that homogeneous blocks are structurally compatible, such that the weights can share with each other. 
During implementation, for each layer, we need to store only the weights of the largest block among the $n$ homogeneous candidates. The remaining smaller building blocks can directly extract weights from the largest one.

Equipped with weight entanglement, one-shot NAS is capable of searching transformer architectures in an efficient and effective fashion, as demonstrated in Fig.~\ref{Fig:SPOS training}. 

Compared with classical weight sharing methods, our weight entanglement strategy has three advantages. 1) \textit{Faster convergence}. Weight entanglement allows each block to be updated more times than the previous independent training strategy. 2) \textit{Low memory cost}. We now only need to store the largest building blocks' parameters for each layer, instead of all the candidates in the space. 3) \textit{Better subnets performance}. We found that the subnets trained with weight entanglement could achieve performance on par with those of training from scratch. 

\subsection{Search Space}
\label{search space}
We design a large transformer search space that includes five variable factors in transformer building blocks: embedding dimension, $Q$-$K$-$V$ dimension, number of heads, MLP ratio, and network depth, as detailed in Tab.~\ref{tab:Search Space} and Fig.~\ref{fig:overview}. 
All these factors are important for model capacities. For example, in attention layers, different heads are used to capture various dependencies. However, recent works \cite{sixheads, voita2019analyzing, bertlook} show that many heads are redundant.
We thereby make the attention head number elastic so that each attention module can decide its necessary number of heads. On the other hand, since different layers have different capacities on feature representation, the varying hidden dimensions in layers might be better than the fixed sizes when used for constructing new models. Moreover, AutoFormer adds new $Q$-$K$-$V$ dimension into the search space and fixes the ratio of the $Q$-$K$-$V$ dimension to the number of heads in each block. This setting makes the scaling factor $\frac{1}{\sqrt{d_h}}$ in attention calculation invariant to the number of heads, stabilizing the gradients, and decouples the meaning of different heads. Besides, we use MLP ratio and embedding dimension together to decide the hidden dimension in each block, which enlarges the search space than using fixed values. 

Following one-shot NAS methods, we encode the search space into a supernet. 
That is, every model in the space is a part/subset of the supernet.
All subnets share the weights of their common parts. The supernet is the largest model in the space, and its architecture is shown in Fig.~\ref{fig:overview}.
In particular, the supernet stacks the maximum number of transformer blocks with the largest embedding dimension, $Q$-$K$-$V$ dimension and MLP ratio as defined in the space.
During training, all possible subnets are uniformly sampled, and the corresponding weights are updated. 

According to the constraints on model parameters, we partition the large-scale search space in to three parts and encode them into three independent supernets, as elaborated in Tab.~\ref{tab:Search Space}. 
Such partition allows the search algorithm to concentrate on finding models within a specific parameter range, which can be specialized by users according to their available resources and application requirements. 

Overall, our supernets contains more than $1.7 \times 10^{16}$ candidate architectures  covering a wide range of model size.

\begin{table}[!t]
	\centering
	\small
	\resizebox{\columnwidth}{!}
	{
		\begin{tabular}{l|ccc} 
			\toprule[1pt]
			&  Supernet-tiny & Supernet-small &  Supernet-base \\
			\midrule[1pt]
			Embed Dim & (192, 240, 24) & (320, 448, 64)  & (528, 624, 48)\\ 
			$Q$-$K$-$V$ Dim & (192, 256, 64) & (320, 448, 64) & (512, 640, 64) \\ 
			MLP Ratio  & (3.5, 4, 0.5)  & (3, 4, 0.5)  & (3, 4, 0.5)  \\
			Head Num & (3, 4, 1) & (5, 7, 1) & (8, 10, 1)\\ 
			Depth Num & (12, 14, 1) & (12, 14, 1) & (14, 16, 1) \\ 
			\midrule[1pt]
			Params Range & 4 -- 9M & 14 -- 34M & 42 -- 75M\\ 
			\bottomrule[1pt]
		\end{tabular} 
	}
		\vspace{1mm}
	\caption{The search space of AutoFormer. We set up three supernets to satisfy different resource constraints. Tuples of three values in parentheses represent the lowest
		value, highest, and steps. \textbf{Note:} the $Q$-$K$-$V$ dimensions, numbers of head  and MLP ratios are varied across layers.}
	\label{tab:Search Space}
	\vspace{-5mm}

\end{table}

\subsection{Search Pipeline}
\label{pipeline}

Our search pipeline includes two sequential phases.

\textit{Phase 1: Supernet Training with Weight Entanglement.}
In each training iteration, we uniformly sample a subnet $\alpha = (\alpha^{(1)}, ... \alpha^{(i)}, ... \alpha^{(l)})$ from the per-defined search space and update its corresponding weights $w = (w^{(1)}, ... w^{(i)}, ... w^{(l)})$ in the supernet's weight $W_\mathcal{A}$ while freezing the rest. Detailed algorithm is given in supplementary materials, Appendix A. 

\textit{Phase 2: Evolution Search under Resource Constraints.}
After obtaining the trained supernet, we perform an evolution search on it to obtain the optimal subnets. Subnets are evaluated and picked according to the manager of the evolution algorithm. Our objective here is to maximize the classification accuracy while minimizing the model size. At the beginning of the evolution search, we pick $N$ random architecture as seeds. The top $k$ architectures are picked as parents to generate the next generation by crossover and mutation. For a crossover, two randomly selected candidates are picked and crossed to produce a new one during each generation. 
For mutation, a candidate mutates its depth with probability $P_d$ first. Then it mutates each block with a probability of $P_m$ to produce a new architecture.

\begin{table*}[!t]
\begin{minipage}[b]{0.55\linewidth}
\centering
	\resizebox{\columnwidth}{!}
{
		\begin{tabular}{l|c c | c } 
			\toprule[1pt]
			Search Method & Inherited & Retrain & Params \\
			\midrule[1pt]
			Random Search   & - & 79.4\% & 23.0M  \\ 
			Classical Weight Sharing + Random Search  & 69.7\% & 80.1\%  & 22.9M \\ 
			Weight Entanglement ~~~~~~+ Random Search    & 81.3\% & 81.4\%  & 22.8M \\ 
			Classical Weight Sharing + Evolution Search (SPOS\cite{spos}) & 71.5\%   & 80.4\% & 22.9M \\ 
			Weight Entanglement ~~~~~~+ Evolution Search (Ours)    &81.7\%   & 81.7\% & 22.9M \\ 
			\bottomrule[1pt]
		\end{tabular} 
    }
    \vspace{5mm}
        \caption{Comparison of different search methods. The supernets are trained for 500 epochs, while the subnets are retrained for 300 epochs. Random search are performed three times and the best performance is reported.}
    \label{tab:Effctiveness}
\end{minipage}\hfill
\begin{minipage}[b]{0.43\linewidth}
\centering
\includegraphics[width=\textwidth]{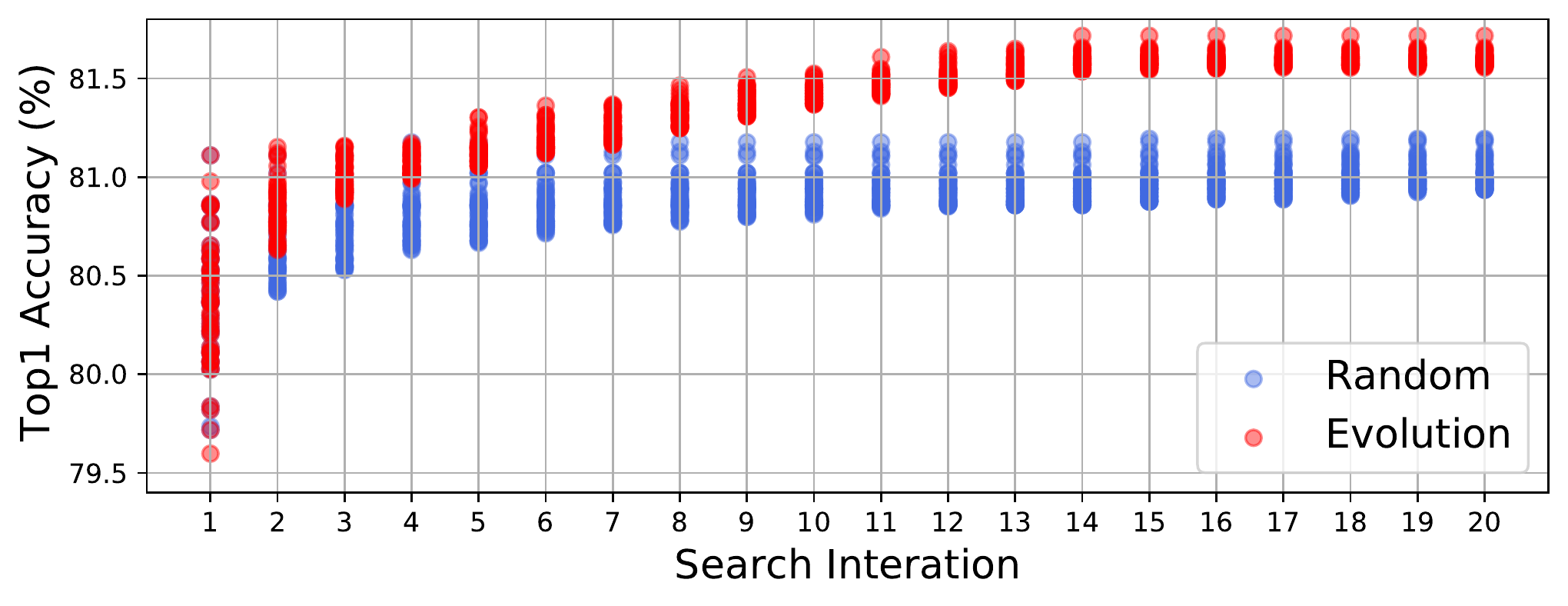}
\captionof{figure}{The performance of subnets inheriting weights from supernet during search. Top 50 candidates until the current iteration are depicted at each search iteration.}
\label{fig:random_vs_evo}
\end{minipage}
\end{table*}

\subsection{Discussion}
\label{discussion}
\textit{Why does weight entanglement work?} 
We conjecture that there are two underlying reasons. 1) Regularization in training. Different from convolution neural networks, transformer has no convolution operations at all. Its two basic components, MSA and MLP, employ only fully connected layers. Weight entanglement could be viewed a regularization training strategy for transformer, to some extent, similar to the effects of dropout \cite{dropout, dropconnect, li2020block}. When sampling the small subnets, corresponding units cannot rely on other hidden units for classification, which hence reduces the reliance of units. 2) Optimization of deep thin subnets. Recent works \cite{bapna2018training, wang2019learning} show that deep transformer is hard to train, which coincides with our observation in Fig.~\ref{fig:motivation}. This is because the gradients might explode or vanish in deep thin networks during backpropagation. Increasing the width of or “overparameterizing” the network will help the optimization \cite{li2017visualizing, du2018gradient, allen2019convergence, zhou2020go}. Our weight entanglement training strategy helps to optimize the thin subnets in a similar way. The gradients backwarded by wide subnets will help to update the weights of thin subnets. Besides, the elastic depth severs 
similar effects to stochastics depth \cite{stochasticdepth} and deep supervision \cite{deepsupervised}, which supervise the shallow layers as well. 

\section{Experiments}

In this section, we first present the implementation details and evolution search settings. We then analyze the proposed weight entanglement strategy and provide a large number of well-trained subnets sampled from supernets to demonstrate its efficacy. At last, we present the performance of AutoFormer evaluated on several benchmarks with comparisons with state-of-the-art models designed manually or automatically.

\subsection{Implementation Details}

\textit{Supernet Training}. We train the supernets using a similar recipe as DeiT \cite{DeiT}. The details are presented in Tab.~\ref{tab: Training settings}. Data augmentation techniques, including RandAugment \cite{cubuk2020randaugment}, Cutmix \cite{yun2019cutmix}, Mixup \cite{zhang2017mixup} and random erasing, are adopted with the same hyperparameters as in DeiT \cite{DeiT} except the repeated augmentation \cite{hoffer2019augment}. Images are split into patches of size 16x16.
All the models are implemented using PyTorch 1.7 and trained on Nvidia Tesla V100 GPUs.

\begin{table}[!t]
	\vspace{-2mm}

	\centering
	\footnotesize
	
	{
		\begin{tabular}{ccccc} 
			\toprule[1pt]
			Epochs & Optimizer & Batch Size & LR & LR scheduler \\
			\midrule[.5pt]
			500 & AdamW & 1024 & 1{\emph e}-3 & cosine \\
			\midrule[1pt]
			Weight & Warmup  & Label  & Drop  & Repeated \\
			Decay & Epochs & Smoothing   &  Path & Augmentation \\
			\midrule[.5pt]
			5{\emph e}-2 & 20 & 0.1 & 0.1 & \ding{55}\\
			\bottomrule[1pt]
		\end{tabular} 
	}
	\vspace{2mm}
	\caption{Supernet training settings. LR refers to learning rate. 
	}
	\label{tab: Training settings}
    \vspace{-1mm}

\end{table}
\begin{table}[!t]

	\centering
	\resizebox{\columnwidth}{!}
    {
		\begin{tabular}{c|c c c c} 
			\toprule[1pt]
			Model & Model Size & Inherited  & Finetune & Retrain \\
			\midrule[1pt]
		    AutoFormer-T  & 5.7M & 74.7\%  & 74.9\%  & 74.9\% \\
		    AutoFormer-S  & 22.9M & 81.7\%  & 81.8\%  & 81.7\% \\ 
		    AutoFormer-B  & 53.7M & 82.4\%  & 82.6\%  & 82.6\% \\ 

		    \bottomrule[1pt]
		\end{tabular} 
	}
	\vspace{0.5mm}
	\caption{Comparison of subnets with inherited weights, fine-tuned (40 epochs) and trained from-scratch (300 epochs).}
	\label{tab:subtransformer}
    \vspace{-3mm}
\end{table}

\textit{Evolutionary Search.} The implementation of evolution search follows the same protocol as in SPOS \cite{spos}. 
For a fair comparison, we reserve the ImageNet validation set for \textit{testing} and subsample 10,000 training examples (100 images per class) as the validation dataset. 
We set the population size to 50 and number of generations to 20. Each generation we pick the top 10 architectures as the parents to generate child networks by mutation and crossover. The mutation probability $P_d$ and $P_m$ are set to 0.2 and 0.4.   

\subsection{Ablation Study and Analysis}

\begin{table*}[!t]
	\caption{AutoFormer performance on ImageNet with comparisons to state-of-the-arts. We group the models according to their parameter sizes. Our AutoFormer consistently outperforms existing transformer-based visual models, being comparable to CNN models.$\dagger$: reported by \cite{DeiT}, $\star$: reported by \cite{xu2021coscale}.}
	\vspace{0.1cm}
	\centering
	\small
    {%
		\begin{tabular}{cc||cc||ccc||cc} 
			\toprule[1.5pt]
			\multicolumn{2}{c||}{Models} & Top-1 Acc. & Top-5 Acc. &\#Parameters & FLOPs  & Resolution & Model Type & Design Type \\
			
			\midrule[1.5pt]

		    \multicolumn{2}{l||}{$\text{MobileNetV3}_{Large1.0}$ \cite{mobilenetv3}} & 75.2\% & - & 5.4M & 0.22G  &  $224^2$ & CNN & Auto \\
		    \multicolumn{2}{l||}{EfficietNet-B0\cite{EfficientNet}} & 77.1\% & 93.3\% & 5.4M & 0.39G &  $224^2$ & CNN & Auto \\
		    \multicolumn{2}{l||}{DeiT-tiny \cite{DeiT}} & 72.2\% & 91.1\% & 5.7M & 1.2G & $224^2$ & Transformer & Manual \\
		    \multicolumn{2}{l||}{\textbf{AutoFormer-tiny (Ours)}} & \textbf{74.7\%} & \textbf{92.6\%} &  \textbf{5.7M} & \textbf{1.3G}  & $224^2$ & Transformer & Auto \\

			\midrule[1.0pt]	

		    \multicolumn{2}{l||}{$\text{ResNet50}^\star$ \cite{he2016deep}} & 79.1\% & - & 25.5M & 4.1G &  $224^2$ & CNN & Manual\\
		    \multicolumn{2}{l||}{$\text{RegNetY-4GF}^\dagger$ \cite{radosavovic2020designing}} & 80.0\% & - & 21.4M & 4.0G  & $224^2$ & CNN & Auto \\
		    \multicolumn{2}{l||}{EfficietNet-B4 \cite{EfficientNet}} & 82.9\% & 95.7\% & 19.3M & 4.2G & $380^2$ & CNN & Auto \\
		    \multicolumn{2}{l||}{BoTNet-S1-59 \cite{srinivas2021bottleneck}} & 81.7\% & 95.8\% & 33.5M  & 7.3G &  $224^2$ & CNN + Trans & Manual\\
		    \multicolumn{2}{l||}{T2T-ViT-14 \cite{T2TViT}} & 81.7\% & - & 21.5M  & 6.1G &  $224^2$ & Transformer & Manual\\
		    \multicolumn{2}{l||}{DeiT-S \cite{DeiT}} & 79.9\% & 95.0\% & 22.1M & 4.7G &  $224^2$ & Transformer & Manual\\
		    \multicolumn{2}{l||}{ViT-S/16 \cite{dosovitskiy2020vit}} & 78.8\% & - & 22.1M  & 4.7G &  $384^2$ & Transformer & Manual\\

			\multicolumn{2}{l||}{\textbf{AutoFormer-small (Ours)}} & \textbf{81.7\% }& \textbf{95.7\%} & \textbf{22.9M} & \textbf{5.1G}  &  $224^2$ & Transformer & Auto\\

			\midrule[1.0pt] 

		    \multicolumn{2}{l||}{$\text{ResNet152}^\star$ \cite{he2016deep}} & 80.8\% & - & 60M & 11G  &  $224^2$ & CNN & Manual \\
		    \multicolumn{2}{l||}{EfficietNet-B7 \cite{EfficientNet}} & 84.3\% & 97.0\% & 66M & 37G  &  $600^2$ & CNN & Auto \\
		    \multicolumn{2}{l||}{ViT-B/16 \cite{dosovitskiy2020vit}} & 79.7\% & - & 86M & 18G &  $384^2$ & Transformer & Manual \\
		    \multicolumn{2}{l||}{Deit-B \cite{DeiT}} & 81.8\% & 95.6\% & 86M & 18G  &  $224^2$ & Transformer & Manual  \\
		    \multicolumn{2}{l||}{\textbf{AutoFormer-base (Ours)}} & \textbf{82.4\%}& \textbf{95.7\%} & \textbf{54M} & \textbf{11G}  &  $224^2$ & Transformer & Auto\\

		    \bottomrule[1.5pt]

		\end{tabular} 
	}
	\vspace{-0.2cm}
	\label{tab:sota}
\end{table*}

\textit{The Efficacy of Weight Entanglement. }
We compare AutoFormer with random search and SPOS \cite{spos} (classical weight sharing) baselines to demonstrate the effectiveness of weight entanglement. For random search, we randomly pick up architectures from the search space to meet the model size constraints. For SPOS \cite{spos}, we adapt it to the transformer search space defined in Tab.~\ref{tab:Search Space} and keep the remaining configurations to be consistent with the original method. 
In other words, in each layer of the SPOS supernet, the transformer blocks with different architecture arguments do not share weights. For example, in an MLP block, there are multiple choices of hidden dimensions. Each MLP choice has its own weights, being independent of each other. 
After training the SPOS supernet, we apply the same evolution process to find the most promising architecture candidate and retrain it using the same setting as our AutoFormer.

\begin{figure}[t!]
    \vspace{-5mm}

    \centering
    \includegraphics[width=0.95\columnwidth]{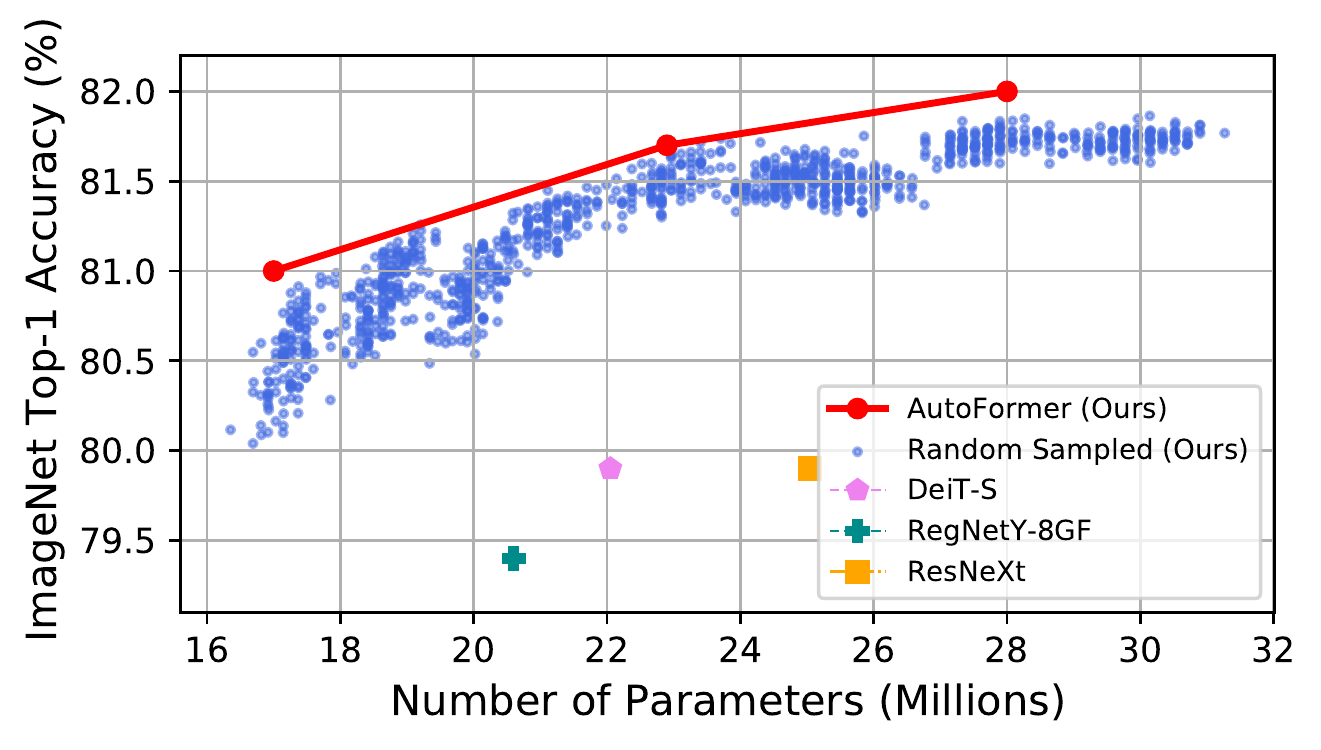}
    \caption{Top-1 accuracy on ImageNet of AutoFormer and 1000 sampled high-performing architectures from the supernet-small with weight inherited from the supernet.}
    \label{fig:many points}
    \vspace{-5mm}

\end{figure}
Table~\ref{tab:Effctiveness} presents the comparisons on ImageNet. We can observe that: 
1) After retraining,random search and SPOS are 2.3\% and 1.3\% inferior to our methods indicating the superiority of the proposed methods.
2) Without retraining, \emph{i.e.}, inheriting weights directly from the supernet, the weight entanglement training strategy can produce significantly better results than the classical weight sharing. The entangled weight can produce well-trained subnets, which are very close to the ones retrained from scratch. We conjecture the relatively inferior performance of SPOS in transformer space is mainly due to insufficient training. We also observe that if we train the supernet in SPOS for more epochs, the performance can be slowly improved. However, its training cost is largely higher than our proposed weight entanglement strategy. 
Fig.~\ref{fig:random_vs_evo} plots the accuracy over the number of architectures sampled from the trained supernet during search. Top 50 candidates are depicted at each generation. It is clear that the evolution search on the supernet is more effective than the random search baseline.

\definecolor{brilliantrose}{rgb}{1.0, 0.33, 0.64}
\textit{Subnet Performance without Retraining}. We surprisingly observe that \emph{there are a large number of subnets achieving very good performance when inheriting weights from the supernets, without extra finetuning or retraining}. The \textcolor{blue}{blue} points shown in Fig.~\ref{fig:many points} represents the 1000 high-performing subsets sampled from the supernet-S. All these subsets can achieve top-1 accuracies ranging from 80.1\% to 82.0\%, exceeding the recent DeiT \cite{DeiT} and RegNetY \cite{radosavovic2020designing}. Such results amply demonstrate the effectiveness of the proposed weight entanglement strategy for one-shot supernet training. 
Tab.~\ref{tab:subtransformer} shows that if we further finetune or retrain the searched subnets on ImageNet, the performance gains are very small, even negligible. 
This phenomenon illustrates the weight entanglement strategy allows the subsets to be well-trained in supernets, leading to the facts that searched transformers do not require any retraining or finetuning and the supernet itself serves good indicator of subnets' ranking.

\begin{table*}[tp]
	\caption{AutoFormer results on downstream classification datasets. $\uparrow$ 384 denotes fine-tuning with 384×384 resolution.}
    \small
    \centering
	\resizebox{\textwidth}{!}
		{
		\begin{tabular}{l|cc|c|ccccc|cc}
				\toprule[1pt]
				Model & \#Param & FLOPs & ImageNet  & CIFAR-10 & CIFAR-100 & Flowers & Cars & Pets & Model Type & Design Type\\
				\midrule[0.75pt]
				$\text{Grafit ResNet-50}$ \cite{touvron2020grafit}& 25M & 12.1G & 79.6 & - & - & 98.2 & 92.5 & - & CNN & Manual\\
				$\text{Grafit RegNetY-8GF}$ \cite{touvron2020grafit}& 39M & 23.4G & 79.6  & - & - & 99.0 & 94.0 & - & CNN & Manual\\
				$\text{EfficientNet-B5}$ \cite{EfficientNet}& 30M & 9.5G & 83.6 & 98.7 & 91.1 & 98.5 & - & - & CNN & Auto\\
				\midrule[0.75pt]
				$\text{ViT-B/16}$  \cite{dosovitskiy2020vit} & 86M & 55.4G & 77.9 & 98.1 & 87.1 & 89.5 & - & 93.8 & Trans & Manual\\
				$\text{DeiT-B $\uparrow$ 384}$ \cite{DeiT}& 86M & 55.4G & 83.1 & 99.1 & 90.8 &  98.5 & 93.3 & - & Trans & Manual\\
				$\textbf{AutoFormer-S $\uparrow$ 384}$ & 23M & 16.5G & 83.4 & 99.1 & 91.1 & 98.8 & 93.4 & 94.9 & Trans & Auto \\

				\bottomrule[1pt]
			\end{tabular}
		}
	\label{tab:transfer}
\vspace{-0.2cm}
\end{table*}
\subsection{Results on ImageNet}
We perform the search of AutoFormer on ImageNet and find multiple transformer models with diverse parameter sizes. All these models inherit weights from supernets directly, without extra retraining and other postprocessing. The performance are reported in Tab.~\ref{tab:sota} and Fig.~\ref{fig:acc}. It is clear that our AutoFormer model families achieve higher accuracies than the recent handcrafted state-of-the-art transformer models such as ViT~\cite{dosovitskiy2020vit} and DeiT \cite{DeiT}. In particular, using $\sim$23M parameters, our small model, \emph{i.e.} AutoFormer-S, achieves a top-1 accuracy of 81.7\%, being 1.8\% and 2.9\% better than DeiT-S and ViT-S/16, respectively. 

Compared to vanilla CNN models, AutoFormer is also competitive. As visualized in Fig.~\ref{fig:acc}, our AutoFormers perform better than the manually-designed ResNet \cite{he2016deep}, ResNeXt \cite{xie2017aggregated} and DenseNet \cite{huang2017densely}, demonstrating the potentials of pure transformer models for visual representation.

However, transformer-based vision models, including AutoFormer, now are still inferior to the models based on inverted residual blocks \cite{sandler2018mobilenetv2}, such as MobileNetV3 \cite{mobilenetv3} and EfficientNet \cite{EfficientNet}. The reason is that inverted residuals are optimized for edge devices, so the model sizes and FLOPs are much smaller than vision transformers.

\subsection{Transfer Learning Results}

\textbf{Classification.} We transfer Autoformer to a list of commonly used recognition datasets: 1) general classification: CIFAR-10 and CIFAR-100 \cite{cifar}; 2) fine-grained classification: Stanford Car~\cite{cars}, FLowers~\cite{flowers} and Oxford-III Pets~\cite{pets}. We follow the same training settings as DeiT~\cite{DeiT}, which take ImageNet pretrained checkpoints and finetune on new datasets.
Tab.~\ref{tab:transfer} shows the results in terms of top-1 accuracy: 1) Compared to state-of-the-art ConvNets, AutoFormer is close to the best results with a negligible gap with fewer parameters; 2) Compared to transformer-based models, AutoFormer achieves better or comparable results on all datasets, with much fewer parameters ($\sim$4x). In general, our AutoFormer consistently achieve comparable or better accuracy with an order of magnitude fewer parameters than existing models, including ResNet~\cite{touvron2020grafit}, RegNet~\cite{touvron2020grafit}, EfficientNet~\cite{EfficientNet}, ViT~\cite{dosovitskiy2020vit}, and DeiT~\cite{DeiT} .

\textbf{Distillation.} AutoFormer is also orthogonal to knowledge distillation (KD) since we focus on searching for an efficient architecture while KD focuses on better training a given architecture. Combining KD with AutoFormers by distilling hard labels from a RegNetY-32GF \cite{radosavovic2020designing} teacher could further improve the performance from 74.7\%/81.7\%/82.4\% to 75.7\%/82.4\%/82.9\%, respectively.

\section{Related Work}

\textit{Vision Transformer.}
Transformer is originally proposed for language modeling \cite{vaswani2017attention}, and recently applied in computer vision. It has shown promising potentials on a variety of tasks, such as recognition, detection and segmentation~\cite{DETR, dosovitskiy2020vit, liang2020polytransform}. A straightforward approach for using transformer in vision is to combine convolutional layers with the self-attention module~\cite{vaswani2017attention,wang2018non}. There has been progress in this direction in recent works, such as~\cite{ramachandran2019stand, zhao2020exploring, wang2020axial,huang2019ccnet}.

Most recently, Dosovitskiy \emph{et al.} introduce Vision Transformer (ViT) \cite{dosovitskiy2020vit}, a pure transformer architecture for visual recognition, without using any convolutions. 
ViT stacks transformer blocks on linear projections of non-overlapping image patches. 
It presents promising results when trained with an extensive image dataset (JFT-300M, 300 million images) that is not publicly available. The most recent DeiT~\cite{DeiT, T2TViT} models verify that large-scale data is not required. 
Using only Imagenet can also produce a competitive convolution-free transformer. 
However, existing visions of transformer models are all built upon manual design, which is engineering-expensive and error-prone. In this work, we present the first effort on automating the design of vision transformer with neural architecture search.

\textit{Neural Architecture Search.} There has been an increasing interest in NAS for automating network design~\cite{Survey,Survey2}. Early approaches search a network using either reinforcement learning~\cite{NASNet,zoph2016neural} or evolution algorithms~\cite{xie2017genetic,real2019regularized}. Most recent works resort to the one-shot weight sharing strategy to amortize the searching cost~\cite{randomNAS,ENAS,SMASH,spos}. 
The key idea is to train a over-parameterized supernet model, and then share the weights across subnets. Among them, SPOS is simple and representative~\cite{spos}. In each iteration, it only samples one random path and trains the path using one batch data.  
Once the training process is finished, the subnets can be ranked by inheriting the shared weights. However, most weight-sharing methods need an additional \emph{retraining} step after the best architecture is identified \cite{spos, blockwise, wu2019fbnet}.

Recent works, OFA~\cite{ofa}, BigNAS \cite{bignas} and slimmable networks \cite{yu2018slimmable, yu2019universally} alleviate this issue by training a once-for-all supernet. Despite the fact that AutoFormer shares similarities with these methods in the concept of training a one-for-all supernet, these methods are designed to search for convolutional networks rather than vision transformers, which have very different architecture characteristics. Specifically, AutoFormer considers the design of multihead self-attention and MLP, which is unique to transformer models, and gives dedicated design of search dimensions as elaborated in Sec.~\ref{search space}. Moreover, BigNAS adopts several well-crafted techniques, such as sandwich training, inplace distillation, regularization, \textit{etc}. OFA proposes a progressively shrinking approach by progressively distilling the full network to obtain the smaller subnets.  By contrast, AutoFormer is simple and efficient, achieving once-for-all training without these techniques, and could be easily extended to search for other vision transformer variants. 

For transformers, there are few studies applying NAS to improve their architectures~\cite{EvolvedTrans,HAT}. These approaches mainly focus on natural language processing tasks. Among them, the most similar one to us is HAT \cite{HAT}. In addition to the difference between tasks, HAT requires an additional \emph{retraining} or \emph{finetuning} step after the search, while AutoFormer does not, which is the key difference. Another difference is the search space. HAT searches for an encoder-decoder Transformer structure, while ours is a pure encoder one. There are two concurrent works, \emph{i.e.}, BossNAS \cite{li2021bossnas} and CvT \cite{wu2021cvt}, exploring different search space from ours. BossNAS searches for CNN-transformer hybrids, while ours for pure transformers. CvT proposes a new architecture family and searches for the strides and kernel size of them. Due to the difference of search space, we do not compare them in this work.

In summary, our work differs from previous works in two main aspects. First, we propose a simple and efficient search methodology dedicated to the structure of vision transformers, which has not been well studied.
Second, the proposed weight entanglement training strategy enables the subnets in the supernet to be well trained, such that the subnets can inherit weights from the one-shot supernet directly, without extra finetuning or retraining. 
To our best knowledge, this is the first work to train a once-for-all transformer in architecture search.

\section{Conclusion}
In this work, we propose a new one-shot architecture search method, AutoFormer, dedicated to transformer search. AutoFormer is equipped with the training strategy, \textit{Weight Entanglement}. Under this strategy, the subnets in the search space are almost fully trained. Extensive experiments demonstrate the proposed algorithm can improve the training of supernet and find promising architectures. Our searched AutoFormers achieve state-of-the-art results on ImageNet among vision transformers. Moreover, AutoFormers transfer well to several downstream classification tasks and could be further improved by distillation. In future work, we are interested in further enriching the search space by including convolutions as new candidate operators. Applying weight entanglement to convolution network search or giving the theoretical analysis of the weight entanglement are other potential research directions.

{\small
\bibliographystyle{ieee_fullname}
\bibliography{egbib}
}

\end{document}